# Axially-shifted pattern illumination for macroscale turbidity suppression and virtual volumetric confocal imaging without axial scanning

SHAOWEI JIANG,[1] JUN LIAO,[1] ZICHAO BIAN,[1] PENGMING SONG,[2] GARRETT SOLER,[1] KAZUNORI HOSHINO,[1] AND GUOAN ZHENG,[1,2,*]

[1]*Department of Biomedical Engineering, University of Connecticut, Storrs, CT, 06269, USA*
[2]*Department of Electrical and Computer Engineering, University of Connecticut, Storrs, CT, 06269, USA*
*\*Corresponding author: guoan.zheng@uconn.edu*

**Structured illumination has been widely used for optical sectioning and 3D surface recovery. In a typical implementation, multiple images under non-uniform pattern illumination are used to recover a single object section. Axial scanning of the sample or the objective lens is needed for acquiring the 3D volumetric data. Here we demonstrate the use of axially-shifted pattern illumination (asPI) for virtual volumetric confocal imaging without axial scanning. In the reported approach, we project illumination patterns at a tilted angle with respect to the detection optics. As such, the illumination patterns shift laterally at different z sections and the sample information at different z-sections can be recovered based on the captured 2D images. We demonstrate the reported approach for virtual confocal imaging through a diffusing layer and underwater 3D imaging through diluted milk. We show that we can acquire the entire confocal volume in ~1s with a throughput of 420 megapixels per second. Our approach may provide new insights for developing confocal light ranging and detection systems in degraded visual environments.**

*OCIS codes: (110.0113) Imaging through turbid media; (180.0180) Confocal microscopy; (150.6910) Three-dimensional sensing*

Structured illumination has been widely used in optical sectioning [1-4] and 3D surface measurement [5, 6]. In a typical sectioning implementation, a sinusoidal illumination pattern is projected onto the specimen and laterally translated to three different positions by one-third of the period. The resulting image set contains the sum of contributions from the in-focus plane, which are modulated by the sinusoidal pattern, plus out-of-focus planes (not focused) that do not contain the distinct sinusoidal pattern from the illumination. The acquired imaging set is then treated with a simple algebraic function to produce a single in-focus section that is free of the blur arising from the remote focal planes. Axial scanning of the sample or the objective lens is needed for acquiring the 3D volumetric data in such implementations.

For 3D surface measurement, the structured illumination technique uses a projector to produce structured patterns on the object and a camera to acquire the corresponding images. In such a system, the 3D surface map can be recovered by analyzing the distortion of the structured patterns on the object. One prominent example of this technique is the Face ID module in the recent iPhone models, where a dot projector produces a structured illumination pattern on the object and an infrared camera acquires the image of the reflected light. The distorted pattern caused by the object is then analyzed for 3D surface reconstruction. Similarly, a 3D laser light scanner projects a pattern of parallel stripes on the object and recover the 3D surface information based on the geometrical deformation of the strips.

In this letter, we demonstrate the use of axially-shifted pattern illumination (asPI) for virtual volumetric confocal imaging without axial scanning. In our approach, we project slit-array illumination patterns at a tilted angle with respect to the detection optics. The slit-array patterns, therefore, shift laterally at different z sections [7, 8]. The key innovation of the reported approach is to integrate the confocal microscopy concept with the asPI scheme for macroscale turbidity suppression and virtual volumetric confocal imaging. Based on the captured 2D images under the asPI, we recover the 3D volumetric confocal data without axial scanning. The reported approach is an extension of the conventional surface measurement technique for recovering volumetric data instead of the surface map. It also shares its root with digitally scanned light-sheet microscopy [9]. Different from the regular confocal microscopy implementations, the reported approach recovers the volumetric confocal data without involving axial scanning, shortening the acquisition time of 3D semi-transparent objects. Our approach may find applications in stereoscopy, endoscopy, macroscale virtual confocal imaging, and underwater imaging. It may also provide new insights for developing confocal light ranging and detection systems (LiDAR) in degraded visual environments.

The proposed setup is shown in Fig. 1, where we use a digital micro-mirror device (DMD) and lens 1 to project a slit-array illumination pattern [10] onto the 3D object through a diffusing layer. Lens 2 is used to acquire images of the object. The angle between the projection and detection light paths is ~25 degrees. The distance between the object and the lens 2 is ~1 meter.

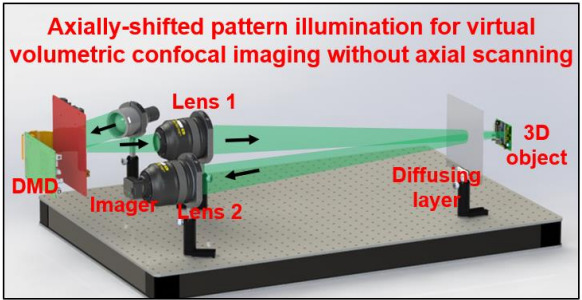

Fig. 1 Rapid virtual volumetric confocal imaging via axially-shifted pattern illumination. Lens 1 (Nikon 50 mm, f / 1.4) and DMD are used to project the axially-shifted illumination patterns onto the object. Lens 2 (Nikon 50 mm, f / 1.4) is attached to an image sensor for image acquisition.

The recovery of one object section at z = $z_1$ can be detailed as follows: 1) we first place a flat reference object (an oxidized aluminum metal plate) at z = $z_1$ and project the slit-array pattern onto the object. 2) The reflected light from the reference object is acquired by the camera in Fig. 2. The acquired pattern is denoted as $M_{x1z1}$, where letter '$M$' means virtual confocal mask [11], the subscript '$z_1$' means the reference pattern is at the $z_1$ position along the z-axis, and the subscript '$x_1$' means the slit-array pattern is at the $x_1$ position along the x-axis (Fig. 2(b1)). 3) We then project a laterally-shifted slit-array pattern to acquire $M_{x2z1}$, where $x_2$ means the slit-array pattern is at the $x_2$ position along the x-axis. In our implementation, we shift the slit-array pattern by 1 DMD pixel each time. If the gap between adjacent slits is 30 pixels on the DMD, we need to acquire 30 confocal masks with one pixel shifted between the adjacent masks: $M_{x1z1}$, $M_{x2z1}$, $M_{x3z1}$, ..., and $M_{x30z1}$ (Visualization 1). We note that, the gap between adjacent slits is determined by the axial range for virtual confocal sectioning. If the gap is denoted as $d$, the achievable axial range will be $d/\tan(\theta)$, where $\theta$ is the angle between the projection and the detection light paths in Fig. 2(a).

With the 30 reference masks captured at the $z_1$ position, we can recover the object section at the $z_1$ position by replacing the flat reference plate with the 3D object we aim to measure. By projecting the same 30 patterns onto the object, we can capture 30 corresponding object images: $O_{x1}$, $O_{x2}$, $O_{x3}$, ..., and $O_{x30}$. The object section at the z1 position, $I_{z1}$, can be recovered by

$$I_{z1} = (O_{x1}M_{x1z1} + ... + O_{x30}M_{x30z1}) / (M_{x1z1} + M_{x2z1} + ... + M_{x30z1}), \quad (1)$$

where the point-wise multiplication between the reference mask and the object image represents the virtual confocal filtering process. This can be explained by the operation of the regular point-scanning confocal microscope, where a focused spot is projected onto the sample and a confocal pinhole is used to reject the object light from other focal planes. In our case, the focused spot is replaced by the slit-array illumination pattern and the physical pinhole is replaced by the multiplication process between the reference mask and the object image. As such, only the region with the slit-array illumination pattern will contribute to the final confocal section. Light from other sections are rejected in the mask-object multiplication process. The lateral slit-array scanning process in our approach is similar to the lateral point scanning process in the regular confocal setup. The summation of the reference masks on the right-hand side of Eq. (1) is used to normalize the non-uniformity of the illumination patterns. The confocal pinhole size in a regular confocal microscope is equivalent to the linewidth of the confocal mask $M$ in our implementation. In a regular confocal implementation, one can shrink the physical confocal pinhole size to get a better sectioning effect. In our implementation, one can also reduce the linewidth of slit-array mask $M$ to achieve the same effect. The extreme case is a slit-array pattern with only one pixel per slit, i.e., to perform thresholding on $M$ to set the maximum intensity columns of the slits to ones and other columns to 0 [3, 11].

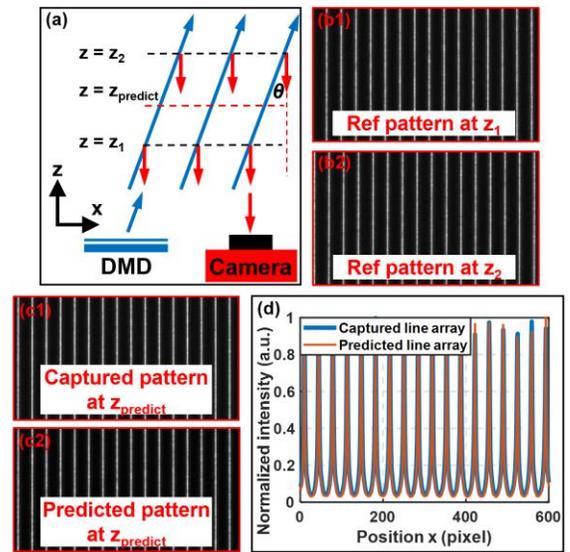

Fig. 2 We use 30 slit-array patterns for sample illumination (Visualization 1). (a) The slit-array pattern shifts laterally as it propagates along the optical axis. (b) The captured reference illumination patterns at two z-positions. These reference patterns serve as virtual confocal masks in the reconstruction process. (c1) The captured reference illumination pattern at z = $z_{predict}$. (c2) We can also infer the reference illumination pattern at $z_{predict}$ via a simple affine transformation. (d) Line traces of (c1) and (c2).

Object sections at other z positions can be recovered in a similar manner, i.e., simply replace $z_1$ in Eq. (1) with other z positions. Reference confocal masks at other z positions are needed in this case. If we want to recover 100 confocal sections ($z_1$, $z_2$, ... $z_{100}$), we need to move the flat reference plate to 100 z-positions, and for each z position, 30 confocal masks will be captured by projecting the laterally-shifted slit-array patterns on the reference flat plate. In practice, we only need to capture 3 reference confocal masks regardless how many confocal sections we want to recover. These 3 masks can be, for example, $M_{x1z1}$, $M_{x30z1}$, $M_{x1z100}$. Other confocal masks can be directly generated through an affine transformation of these three masks. For example, $M_{x1z1}$ and $M_{x30z1}$ can be used to generate an affine map to shift the projected slit-array pattern at the z1 position. Similarly, $M_{x1z1}$ and $M_{x1z100}$ can be used to generate another affine map to shift the projected slit-array pattern to different z sections. Figure 2(b) shows two captured reference confocal masks (i.e., illumination patterns) at the $z_1$ and $z_2$ positions. Based on these two masks, we first generate an affine map and then apply the map to generate a reference confocal mask at the $z_{predict}$ position (Fig. 2(c2)). We compare our generated mask with the captured mask in Fig. 2(c)-(d), and they are in good agreement with each other. There is no observable difference between the use of the captured patterns and the predicted pattern for image reconstruction.

In the first experiment, we demonstrate our approach for imaging a multilayer semi-transparent object (three cover glasses with white paint). Figure 3(a) shows the captured regular 2D image of the three-layer object under uniform illumination. The recovered confocal sections are shown in Fig. 3(b) and the ground truths are shown in Fig. 3(c) for comparison. In this experiment, the gap between the adjacent slits is 60 pixels and we recover 100 z-sections with a 100-μm step size (Visualization 2).

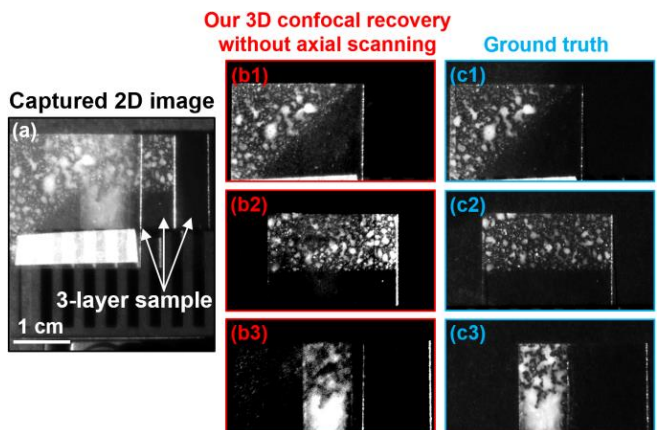

Fig. 3 (Visualization 2) Volumetric confocal imaging of semi-transparent object without axial scanning. (a) The captured regular 2D image via uniform illumination. (b) The recovered confocal sections without axial scanning. (c) The ground truth of the three sections (manually removing two other sections).

In the second experiment, we demonstrate our approach for 3D imaging through a diffusing layer (a circuit board hidden behind a glass plate with white paint). Figure 4(a) shows the captured regular 2D image of circuit board hidden behind the diffusing layer. Figure 4(b1) and 4(b2) show the recovered confocal sections of the diffusing layer and the circuit board. Figure 4(c) shows the 3D recovery of the circuit board through the diffusing layer, without involving axial scanning. In this experiment, the gap between the adjacent slits is 120 pixels and we recover 300 z-sections with a 50-μm step size (Visualization 3).

In the third experiment, we demonstrate our approach for underwater imaging through diluted milk. Figure 5(a) shows the setup with a tilted object (glass plate with white paint) placed in the water tank. Figure 5(b) and 5(c) show the regular 2D images with clear water and with diluted milk. Our 3D confocal recoveries are shown in Fig. 5(d) for clear water and in Fig. 5(e) for diluted milk. In this experiment, the gap between the adjacent slits is 120 pixels and we recover 400 z-sections with a 50-μm step size.

In the fourth experiment, we demonstrate our approach for 3D surface imaging. Figure 6(a) shows the regular 2D images of three different objects. Figure 6(b) shows one confocal section via our approach. For 3D surface recovery, we locate the highest intensity pixel along different z-sections to recover the depth. Figure 6(c) shows the surface recovery of the three objects. In this experiment, the gap between the adjacent slits is 120 pixels and we recover 400 z-sections with a 50-μm step size (Visualizations 4-6).

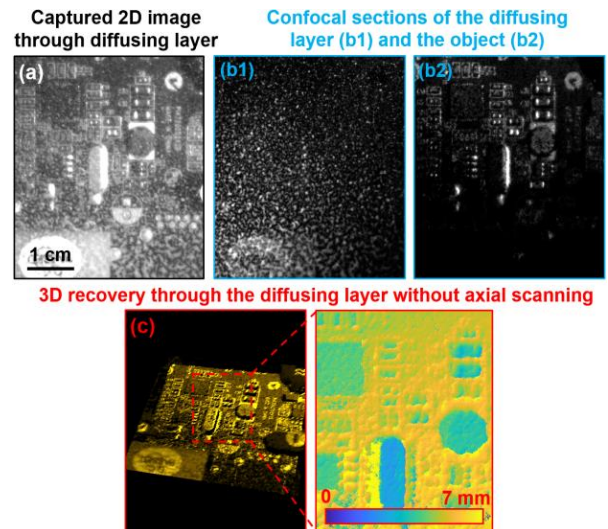

Fig. 4 (Visualization 3) 3D confocal imaging through a diffusing layer (the setup is shown in Fig. 1). The scattering mean free path for the diffusing layer is ~20 μm and the thickness of this layer is ~50 μm. (a) The captured image via uniform illumination. One confocal section of the diffusing layer (b1) and the circuit-board object (b2). (c) The recovered depth map of the circuit-board through the diffusing layer.

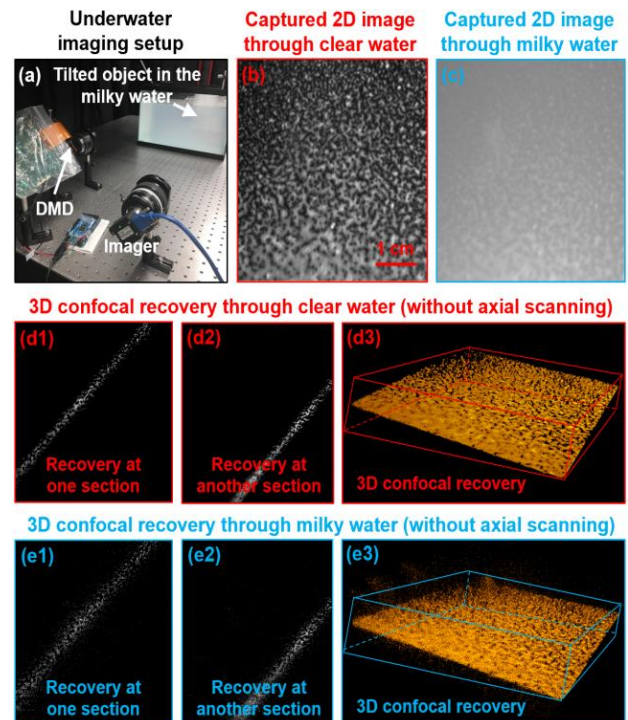

Fig. 5 3D underwater imaging through diluted milk. The scattering mean free path is ~6 cm for the diluted milk and the thickness of the scattering media is ~20 cm. (a) The setup for underwater imaging, where a tilted object is placed in the water tank. The captured images under uniform illumination through clear water (b) and milky water (c). Our 3D confocal recovery of the tilted object with clear water (d) and milky water (e).

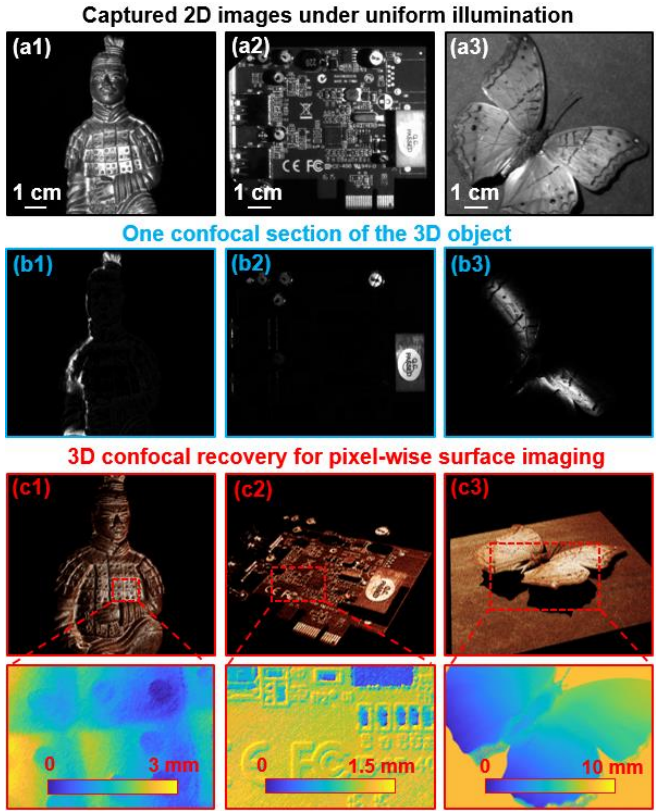

Fig. 6 (Visualizations 4-6) Pixel-wise 3D depth imaging via the reported approach. (a) The captured images via uniform illumination. (b) One confocal section recovered via our approach. (c) The 3D renderings based on the recovered 3D depth maps.

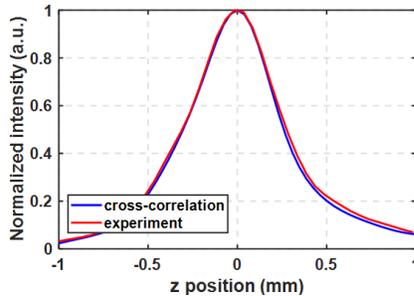

Fig. 7 The axial point spread function of the reported confocal imaging system. The red curve shows the experimental axial response of a single-layer uniform object. The blue curve shows the cross-relation between the captured $O(x,y)$ and the virtual mask $M(x,y)$.

The axial point spread function (PSF) of the reported approach is determined by two factors: 1) the line widths of the projected slit-array pattern, and 2) the angle $\theta$ between the detection and projection light paths. To quantify the axial PSF, we use a single-layer uniform flat plate as the object. If the first captured image under slit-array projection is denoted as $O(x,y)$, we have $O_{xi} = O(x-x_i, y)$ (i = 1,2,3…) for Eq. (1). Similarly, if the first virtual mask is denoted as $M(x,y)$, we have $M_{xizj} = M(x-x_i-z_j tan\theta, y)$ (i, j = 1,2,3…). Based on Eq. (1), we can get the following axial PSF for any given position (x, y):

$$PSF(z_j) = \sum_{i=1}^{n} O(x-x_i,y)M(x-x_i-z_j tan\theta, y), \quad (2)$$

where the axial PSF is, essentially, the cross correlation between the slit-array pattern $O(x,y)$ and the virtual mask $M(x,y)$. In Fig. 7, we plot the experimental axial response of a single-layer uniform object in the red curve, with a full width at half maximum of ~0.5 mm. The blue curve in Fig. 7 is the cross-correlation between the captured $O(x,y)$ and the virtual mask $M(x,y)$. They are in a good agreement with each other, validating our analysis.

In summary, we integrate the confocal microscopy concept with the asPI scheme for macroscale turbidity suppression and virtual volumetric confocal imaging. Conventional confocal microscopy typically requires axial scanning of the sample (or the objective lens) to acquire the 3D volumetric data. The reported asPI scheme is a lateral scanning operation at the illumination end and it converts the axial information to the acquired x-y images. As such, no axial scanning is needed, shortening the acquisition time for 3D semi-transparent objects. In our experiments, we synchronize our camera (Andor Zyla 4.2 PLUS, 4.2 megapixels) with the DMD (TI DLPLCR6500) to acquire the full field-of-view images at 100 fps. The entire confocal volume can be captured in ~1s and the corresponding throughput is 420 megapixels per second. The depth of field of our macroscopic imaging system is much larger than the scale of the 3D objects. Therefore, defocusing is not an issue in our implementation. For high numerical aperture (NA) microscopy, however, the depth of field will be much smaller than the axial scale of the object, resulting in just a few optical sections that could be recovered. To this end, the reported scheme is most efficient when the many optical sections are reconstructed over a relatively long depth of field. For few optical sections in a high-NA setting, the three-phase sectioning approach [1] may be more efficient in terms of imaging speed. The reported approach is an extension of conventional surface measurement techniques for virtual volumetric confocal imaging. It may find applications in stereoscopy, endoscopy, macroscale virtual confocal imaging, and underwater imaging. It may also provide new insights for developing confocal light ranging and detection systems (LiDAR) in degraded visual environments.

**Funding.** This work was in part supported by NSF (1809047, 1555986, 1700941), and NIH (R21EB022378, R03EB022144).